\documentclass{article}
\usepackage{floatrow}
\PassOptionsToPackage{numbers, compress}{natbib}

\usepackage[preprint]{nips_2018}

\usepackage[utf8]{inputenc} 
\usepackage[T1]{fontenc}    
\usepackage{hyperref}       
\usepackage{url}            
\usepackage{booktabs}       
\usepackage{amsfonts}       
\usepackage{nicefrac}       
\usepackage{microtype}      
\usepackage{wrapfig}
\usepackage{graphicx}
\usepackage{bm}
\usepackage{amsmath}
\usepackage{mysymbol}
\usepackage{color}
\usepackage{algorithm}
\usepackage{algpseudocode}

\title{Scalable Centralized Deep Multi-Agent Reinforcement Learning via Policy Gradients}

\author{
  Arbaaz Khan, Clark Zhang, Daniel D. Lee, Vijay Kumar, Alejandro Ribeiro \\
  GRASP Laboratory\\
  University of Pennsylvania\\
}

\begin{document}

\maketitle
 
\begin{abstract}
In this paper, we explore using deep reinforcement learning for problems with multiple agents. Most existing methods for deep multi-agent reinforcement learning consider only a small number of agents. When the number of agents increases, the dimensionality of the input and control spaces increase as well, and these methods do not scale well. To address this, we propose casting the multi-agent reinforcement learning problem as a distributed optimization problem. Our algorithm assumes that for multi-agent settings, policies of individual agents in a given population live close to each other in parameter space and can be approximated by a single policy. With this simple assumption, we show our algorithm to be extremely effective for reinforcement learning in multi-agent settings. We demonstrate its effectiveness against existing comparable approaches on co-operative and competitive tasks. 
\end{abstract}

\section{Introduction}
Leveraging the power of deep neural networks in reinforcement learning (RL) has emerged as a successful approach to designing policies that map sensor inputs to control outputs for complex tasks. These include, but are not limited to, learning to play video games~\cite{dqn,mnih2016asynchronous}, learning complex control policies for robot tasks~\cite{visuomotor} and learning to plan with only sensory information~\cite{pathak2017curiosity,macn,gupta2017cognitive}. While these results are impressive, most of these methods consider only single agent settings. 

In the real world, many applications, especially in fields like robotics and communications, require multiple agents to interact with each other in co-operative or competitive settings. Examples include warehouse management with teams of robots~\cite{enright2011optimization}, multi-robot furniture assembly~\cite{knepper2013ikeabot}, and concurrent control and communication for teams of robots~\cite{2017Transactions_Stephan}. Traditionally, these problems were solved by minimizing a carefully set up optimization problem constrained by robot and environment dynamics. Often, these become intractable when adding simple constraints to the problem or by simply increasing the number of agents~\cite{solovey2016hardness}. In this paper, we attempt to solve multi-agent problems by framing them as multi-agent reinforcement learning (MARL) problems and leverage the power of deep neural networks. In MARL, the environment from the perspective on an agent appears non-stationary. This is because the other agents are also changing their policies (due to learning). Traditional RL paradigms such as Q-learning are ill suited for such non-stationary environments. 

Several recent works have proposed using decentralized actor-centralized critic models ~\cite{foerster2017counterfactual,lowe2017multi}. These have been shown to work well when the number of agents being considered is small. Setting up a large number of actor networks is not computationally resource efficient. Further, the input space of the critic network grows quickly with the number of agents. Also, in decentralized frameworks, every agent must estimate and track the other agents~\cite{da2006dealing,sutton2007role}. Most deep RL algorithms are sample inefficient even with only a single agent. Attempting to learn individual policies for multiple agents in a decentralized framework becomes highly inefficient, as we will demonstrate. Thus, attempting to learn multiple policies with limited interaction using decentralized frameworks is often infeasible. 

Instead, we propose the use of a centralized model. Here, all agents become aware of the actions of other agents, which mitigates the non-stationarity. To use a centralized framework for MARL, one must collect experiences from individual agents and then learn to combine these to output actions for all agents. One option is to use high-capacity models like neural networks to learn policies that can map the joint observations of all agents to the joint actions of all agents. This simple approach works when the number of agents is small but suffers from the curse of dimensionality when the number of agents increases. Another possibility is to learn a policy for one agent and fine tune it across all agents but this also turns out to be impractical. To mitigate the problems of scale and limited interaction, we propose using a distributed optimization framework for the MARL problem. The key idea is to learn one policy for all agents that exhibits emergent behaviors when multiple agents interact. This type of policy has been shown to be used in nature~\cite{potter2013microclimatic} as well as in swarm robotics~\cite{rubenstein2012kilobot}. In this paper, the goal is to learn these policies from raw observations and rewards with reinforcement learning.

\begin{figure}[t!]
\centering
\includegraphics[scale=0.3]{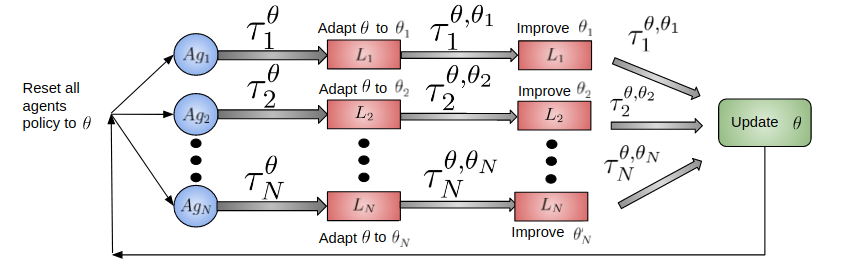}
\caption{\textbf{Multi-Agent framework for Distributed Learning:} Each agent $n$ ($Ag_n$) starts under policy parametrized by $\theta$ and uses it to collect experience $\tau_{n}^{\theta}$. $\tau_{n}^{\theta}$ is used to minimize agent $Ag_{n}$'s loss function $L_n$ and adapt its policy from $\theta$ to $\theta_{n}$. Now, $Ag_n$ uses policy parametrized by $\theta_{n}$ assuming other agents policies remain $\theta$. The trajectory generated in this case is denoted by $\tau_{n}^{\theta,\theta_n}$ and is used to improve $Ag_n$'s policy by taking gradients w.r.t this intermediate policy. Finally, using this improved policy, we collect another new trajectory $\tau_n^{\theta,\theta_n}$. These new trajectories are used to update $\theta$.}
\end{figure}

Optimizing one policy across all agents is difficult and sometimes intractable (especially when number of agents are large). Instead, we take a distributed approach where each agent improves the central policy with their local observations. Then, a central controller combines these improvements in a way that refines the overall policy. This can be seen as recasting the original problem of optimizing one policy to optimizing several policies subject to the constraint that they are identical. After training, there will only be a single policy for all agents to use. This is a optimization technique that has seen success in distributed settings before~\cite{boyd2011distributed}.
Thus the main contributions of this paper are :
\begin{enumerate}
\item A novel algorithm for solving MARL problems using distributed optimization.
\item The policy gradient formulation when using distributed optimization for MARL
\end{enumerate}
\section{Related Work}
Multi-Agent Reinforcement Learning (MARL) has been an actively explored area of research in the field of reinforcement learning~\cite{busoniu2006multi,littman1994markov}. Many initial approaches have been focused on tabular methods to compute Q-values for general sum Markov games~\cite{hu2003nash}. Another approach in the past has been to remove the non-stationarity in MARL by treating each episode as an iterative game, where the other agent is held constant during its turn. In such a game, the proposed algorithm searches for a Nash equilibrium~\cite{conitzer2007awesome}. Naturally, for complex competitive or collaborative tasks with many agents, finding a Nash equilibrium is non-trivial. Building on the recent success of methods for deep RL, there has been a renewed interest in using high capacity models such as neural networks for solving MARL problems. However, this is not very straightforward and is hard to extend to games where the number of agents is more than two~\cite{tampuu2017multiagent}. 

When using deep neural networks for MARL, one method that has worked well in the past is the use of decentralized actors for each agent and a centralized critic with parameter sharing among the agents ~\cite{foerster2017counterfactual,lowe2017multi}. While this works well for a small number of agents, it is sample inefficient and very often, the training becomes unstable when the number of agents in the environment increases. 

In our work, we derive the policy gradient derivation for multiple agents. This derivation is very similar to that for policy gradients in meta-learning from ~\cite{al2017continuous,maml}, where the authors use meta-learning to solve continuous task adaptation. In ~\cite{al2017continuous} the authors propose a meta-learning algorithm that attempts to mitigate the non-stationarity by treating it as a sequence of stationary tasks and train agents to exploit the dependencies between consecutive
tasks such that they can handle similar non stationaries at execution time. This is in contrast to our work where we are focused on the MARL problem. In MARL there are often very few inter-task (in the MARL setting this corresponds to inter-agent) dependencies that can be exploited. Instead, we focus on using distributed learning to learn a policy.

\section{Collaborative Reinforcement Learning in Markov Teams}

We consider policy learning problems in a collaborative Markov team~\cite{littman1994markov}. The team is composed of $N$ agents generically indexed by $n$ which at any given point in time $t$ are described by a state $s_{nt}\in\ccalS$ and an action $a_{nt}\in\ccalA$. Observe that we are assuming all agents to have common state space $\ccalS$ and common action space $\ccalA$. Individual states and actions of each agent are collected in the vectors $s_{t}:=[s_1;\ldots;s_N]\in\ccalS^N$ and $a_{t}:=[a_1;\ldots;a_N]\in\ccalA^N$. Since the {\it team} is assumed to be Markov, the probability distribution of the state at time $t+1$ is completely determined by the conditional transition probability $p \left(s_{t+1}\given s_t, a_t\right)$. We further assume here that agents are statistically identical in that the probability transition kernel is invariant to permutations. 

At any point in time $t$, the agents can communicate their states to each other and agents utilize this information to select their actions. This means that each agent executes a policy $\pi_n:\ccalS^N\to\ccalA$ with the action executed by agent $n$ at time $t$ being $a_{nt} = \pi_{n}(s_t)$. As agents operate in their environment, they collect {\it individual} rewards $r_n(s_t, a_{nt})$ which depend on the state of the team $s_t$ and their own individual action $a_{nt}$. The quantity of interest to agent $n$ is not this instantaneous reward but rather the long term reward accumulated over a time horizon $T$ as discounted by a factor $\gamma$,
\begin{equation}\label{eqn_discounted_reward_individual}
   R_n := \sum_{t=0}^{T} \gamma^t r_n(s_t, a_{nt}).
\end{equation}
The reward $R_n$ in \eqref{eqn_discounted_reward_individual} is stochastic as it depends on the trajectory's realization. In conventional RL problems, agent $n$ would define the cost $\tdL_n(\pi_n) :=\mbE_{\pi_n}(R_n)$ and search for a policy $\pi_n$ that maximizes this long term expected reward. This expectation, however, neglects the effect of other agents, which we can incorporate competitively or collaboratively. In a competitive formulation agent $n$ considers the loss $L_n(\Pi) :=\mbE_{\Pi}(R_n)$ that is integrated not only with respect to its own policy but with respect to the policies of all agents $\Pi := [\pi_1;\ldots;\pi_N]$. In the collaborative problems we consider here, agent $n$ takes the rewards of other agents into consideration. Thus, the reward of interest to agent $n$ is the expected reward accumulated over time and across all agents,
\begin{equation}\label{eqn_joint_reward}
   L(\Pi) =  \mbE_{\Pi}  \bigg[\sum_{n=1}^N R_n \bigg] 
          =  \sum_{n=1}^N \mbE_{\pi_n, \pi_{-n}} [R_n]  
          =  \sum_{n=1}^N L_n(\Pi)
          = \sum_{n=1}^N L_n(\pi_n, \pi_{-n}) .
\end{equation}
where, we recall, $\Pi = [\pi_1;\ldots;\pi_N]$ denotes the joint policy of the team and we have further defined $\pi_{-n} = [\pi_m]_{m\neq n}$ to group the policies of all agents except $n$.

The goal in a collaborative reinforcement learning problem is to find a policies $\pi_n$ that optimize the aggregate expected reward in \eqref{eqn_joint_reward}. We can write these optimal policies as $\Pi^\dagger = \argmax_\Pi (L(\Pi))$. The drawback with this problem formulation is that it requires learning separate policies for each individual agent. This is intractable when $N$ is large, which motivates a restriction in which all agents are required to execute a common policy. This leads to the optimization problem  
\begin{equation}\label{eqn_optimization_premliminary}
   \pi^* :=  \argmax  L(\pi_n, \pi_{-n}), \quad 
             \st      \pi_n = \pi_m, \text{\ for all\ } n\neq m.
\end{equation}
We reformulate into the more tractable problem
\begin{equation}\label{eqn_optimization}
   \pi^* =  \argmax  L (\pi_n, \pi), \quad 
           \st \pi_n = \pi \text{\ for all\ } n
\end{equation}
In the next section, we present a distributed algorithm to solve this optimization problem. 

\section{Distributed Optimization for MARL using Policy Gradients}
Let us reiterate the problem in Eqn \ref{eqn_optimization_premliminary} in terms of the parameterization of the policy and trajectories drawn from the policy. Eqn \ref{eqn_optimization_premliminary} can be interpreted as a problem where we aim to solve is to find the best set of parameters $\theta^*$ that parameterizes a policy $\pi_\theta$ to maximize the sum of rewards $R_i$ for all agents over some time horizon $T$. Specifically, the optimization problem in Eqn \ref{eqn_optimization_premliminary} can be written as:
\begin{equation}\label{opt1}
\theta^* = \max_{\theta} \sum_{n=1}^{N} \mathbb{E}_{\tau_n^{\theta} \sim P_n(\tau_n^{\theta} | \theta)} \Big[ R_n \Big] = \max_{\theta} \sum_{n=1}^{N} L_n(\theta)
\end{equation}
where $\tau_n^{\theta}$ are trajectories of agent $n$
\begin{equation} \label{traj1}
\begin{split}
\tau_{n}^\theta=\Big\{ [s_{n}^{t_0,\theta},a_{n}^{t_0,\theta}, a_{1,\ldots N \neq n}^{t_0,\theta}, r_{n}^{t_0}], [s_{n}^{t_1,\theta},a_{n}^{t_1,\theta}, a_{1,\ldots N \neq n}^{t_1,\theta}, r_{n}^{t_1}]
\ldots,  [s_{n}^{t_T,\theta},a_{n}^{t_T,\theta}, a_{1,\ldots N \neq n}^{t_T,\theta}, r_{n}^{t_T} ] \Big\} 
\end{split}
\end{equation}
sampled from the distribution of trajectories $P_n(\tau_n^{\theta} | \theta)$ induced by the policy $\pi_\theta$. However, as stated above this problem can be intractable for large $N$. Rewriting the parametrized version of the more tractable optimization in Eqn ~\ref{eqn_optimization} we get:
\begin{equation}\label{opt2}
\begin{aligned}
& \underset{\theta, \{\theta_i\}}{\text{max}}
& & \sum_{n=1}^{N} \mathbb{E}_{\tau_n^{\theta,\theta_n} \sim P_n(\tau_n^{\theta,\theta_n} | \theta, \theta_n)} \Big[ R_n \Big] = \max_{\theta} \sum_{n=1}^{N} L_n(\theta, \theta_n)\\
& \text{subject to}
& & \theta_n = \theta, \forall \text{ n}
\end{aligned}
\end{equation}
where we define the trajectories $\tau_n^{\theta,\theta_n}$ to be those obtained when agent $n$ follows policy $\pi_{\theta_n}$ and all other agents follow policy $\pi_\theta$. \footnote{This optimization problem is the same as the one in Eqn ~\ref{eqn_optimization}. The difference being that, we have now written the optimization in terms of the parametrization of the policies and trajectories drawn from the policies.} 
\begin{equation}\label{traj2}
\begin{split}
\tau_{n}^{\theta, \theta_n} =\Big\{ [s_{n}^{t_0,\theta_n},a_{n}^{t_0,\theta_n}, a_{1,\ldots N \neq n}^{t_0,\theta}, r_{n}^{t_0}], [s_{n}^{t_1,\theta_n},a_{n}^{t_1,\theta_n}, a_{1,\ldots N \neq n}^{t_1,\theta}, r_{n}^{t_1}],\\
\ldots, [s_{n}^{t_T,\theta_n},a_{n}^{t_T,\theta_n}, a_{1,\ldots N \neq n}^{t_T,\theta}, r_{n}^{t_T} ] \Big\}
\end{split}
\end{equation}
The difference between Eqn \ref{opt1} and Eqn \ref{opt2} is that we have formed $N$ copies of $\theta$ labeled $\theta_n$ and put a constraint that $\theta = \theta_n$. This approach allows us to look at the problem in a different light. Similar to other distributed optimization problems such as ADMM \cite{boyd2011distributed}, we can decouple the optimization over $\theta_n$ from that of $\theta$. The general approach is an iterative process where
\begin{enumerate}
\item For each agent $n$, optimize the corresponding $\theta_n$ 
\item Consolidate the $\theta_n$ into $\theta$
\end{enumerate}
This is often realized as a projected gradient descent where for each agent $n$, we apply the gradients $\theta_n \leftarrow \theta_n + \alpha_1 \nabla_{\theta_n}L(\theta, \theta_n)$ as well as applying a gradient $\theta \leftarrow \theta + \alpha_2  \nabla_\theta \sum_{n=1}^N L(\theta, \theta_n)$. Then, in the next iteration all agents start at $\theta_n$ where $\theta_n$ is realized by taking a projection step such that $\theta_n = \theta \leftarrow  \frac{1}{N+1} (\theta + \sum_{n=1}^N \theta_n)$ is taken to satisfy the constraint in problem \ref{opt2}. However, when computing this projected gradient step, we need to keep track of all $\theta_n$ to compute the average. This is infeasible if this is done for a large number of agents. Instead a simple approximation to the projected gradient is used  by setting $\theta_n \leftarrow \theta$. In the next subsection, we present our algorithm \textit{Distributed Multi Agent Policy Gradient} or DiMA-PG and its practical implementation.

\subsection{Distributed Multi-Agent Policy Gradients (DIMA-PG)}

In this section, we propose the \textit{Distributed Multi Agent Policy Gradient} (DiMA-PG) algorithm which learns a centralized policy that can be deployed across all agents. Consider a population $Pop$ from which $N$ statistically identical agents are sampled according to a distribution $P(Pop)$. The parameters $\theta_n$ of this agent-specific  policy are updated by taking the gradient w.r.t $\theta$ at the specific value of $\theta = \theta_0$ (where $\theta_0$ is your current central policy):
\begin{equation}
\label{eq:3}
\theta_n \leftarrow \theta_0 + \alpha_1 \nabla_{\theta_n} 
L_n(\theta,\theta_n)|_{\theta=\theta_0, \theta_n=\theta_0}
\end{equation}
where $\alpha$ is step size hyperparameter and 
$L(\theta,\theta_n)$ is as defined in Eqn ~\ref{opt2}. Note that $L(\theta_0, \theta_0)$ uses trajectories $\tau_n^{\theta_0, \theta_0}$ generated when all agents follow policies $\pi_{\theta_0}$ while $L(\theta_0, \theta_n)$ uses trajectories $\tau_n^{\theta_0, \theta_n}$ when agent $n$ follows $\pi_{\theta_n}$ while all other agents follow $\pi_{\theta_0}$.We do this because, when the environment is held constant w.r.t agent, then the problem for agent $n$ reduces to a MDP~\cite{sutton1998reinforcement}. 

In practice, we can take $k$ gradient steps instead of just one as presented in Eqn \ref{eq:3}. This can be done with the following inductive steps
\begin{equation}
\begin{aligned}
\label{eq:multi}
\theta_n^{[0]} &= \theta_0\\
\theta_n^{[k]} &= \theta_n^{[k-1]} + \alpha_2 \nabla_{\theta_n} 
L_n(\theta,\theta_n)|_{\theta=\theta_0, \theta_n=\theta_n^{[k-1]}}\\
\theta_n &= \theta_n^{[k]}
\end{aligned}
\end{equation}
Finally, we update $\theta$: 
\begin{equation}
\label{eq:7}
\theta \leftarrow \theta + \epsilon \nabla_{\theta} \sum_{n=1}^N  L_{n}(\theta, \theta_n)
\end{equation}

Numerically, we approximate $\nabla_{\theta_n} L_n(\theta,\theta_n)$ by drawing $l$ trajectories where agent $n$ uses policy $\pi_{\theta_n}$ while all other agents uses policy $\pi_\theta$ and averaging over the policy gradients ~\cite{reinforce,sutton1998reinforcement} that each trajectory provides. Recall that the trajectories $\tau_{n}^\theta$ and 
$\tau_{n}^{\theta,\theta_n}$ are random variables with distributions $P_{n}(\tau_{n}^\theta|\theta)$ and $P_n(\tau_{n}^{\theta,\theta_n}|\theta, \theta_n)$ respectively. The individual agent policy parameters, $\theta_n$ are also random variables with distribution $P_{n}(\theta_n | \theta)$. The overall optimization can be written as:
\begin{equation}
\label{eq:8}
\max_{\theta} \mathbb{E}_{n \sim P(Pop)} \Big[ \mathbb{E}_{\tau_{n}^\theta \sim P_{n}(\tau^{\theta}_{n}|\theta)} \Big[ \mathbb{E}_{\tau_{n}^{\theta,\theta_n} \sim P_{n}(\tau_{n}^{\theta, \theta_n}|\theta,\theta_n)} [L_{n}(\theta,\theta_n)|(\tau_{n}^{\theta},\theta)] \Big] \Big]
\end{equation}
Assuming, we sample N agents,  Eqn. \ref{eq:8} can be rewritten as: 
\begin{equation}
\label{eq:9}
\max_{\theta} \frac{1}{N}\sum_{n=1}^N \Big[ \mathbb{E}_{\tau_{n}^\theta \sim P_{n}(\tau^{\theta}_{n}|\theta)} \Big[ \mathbb{E}_{\tau_{n}^{\theta,\theta_n} \sim P_{n}(\tau_{n}^{\theta, \theta_n}|\theta,\theta_n)} [L_{n}(\theta,\theta_n)|(\tau_{n}^{\theta},\theta)] \Big] \Big]
\end{equation}
To learn $\theta$, we use policy gradient methods~\cite{reinforce,sutton1998reinforcement} which operate by taking the gradient of Eqn. \ref{eq:9}. One can also use recently proposed state of the art methods for policy gradient methods ~\cite{gae,schulman2015trust}. The gradient for each agent in Eqn ~\ref{eq:9} (the quantity inside the sum) w.r.t $\theta$ can be written as:
\begin{equation}
\label{polgrad}
\begin{split}
\nabla_{\theta} \mathcal{L}_n(\theta,\theta_n) =\mathop{\mathbb{E}}_{\tau_{n}^\theta \sim P_{n}(.|\theta),\tau_{n}^{\theta,\theta_n} \sim P_{n} (.|\theta,\theta_n)} \Bigg[ L_{n}(\theta,\theta_n)\nabla_{\theta} \log \pi_{\theta_{n}}(\tau_{n}^{\theta,\theta_n}) + L_{n}(\theta,\theta_n)\nabla_{\theta}\log \pi_{\theta}(\tau_{n}^{\theta})\Bigg] 
\end{split}
\end{equation}
The policy gradient for each agent consists of two policy gradient terms, one over the trajectories $\tau_{n}^{\theta,\theta_n}$ sampled using ($\theta,\theta_n$) and another term over the trajectories $\tau_{n}^{\theta}$ sampled using $\theta$. It may be noted that the terms from the agent specific policy improvement when the other agents are held stationary (Eqn \ref{eq:multi}) do not appear in the final term. We show that it is possible to marginalize these terms out in the derivation for the gradient and point the reader to the appendix for a full derivation of the policy gradient. The full algorithm for DiMA-PG is presented in Algorithm \ref{algodimapg}. 

\begin{algorithm}
  \caption{\textbf{Distributed Multi Agent with Policy Gradients (DIMA-PG)}}\label{algodimapg}
  \begin{algorithmic}[1]
  	\Require Initial random central policy $\theta$, step-size hyperparameters $\alpha_1,\alpha_2, \epsilon$ and distribution over agent population P(Pop)
      \While{True}
        \State Sample $N$ agents $\sim$ P(Pop)
        \For {all agents}
        \State Collect trajectory $\tau_n^{\theta}$ as given in Eqn ~\ref{traj1} and evaluate agent loss $L_n({\theta,\theta_n})|_{\theta=\theta_0, \theta_n=\theta_0}$
        \State Compute agent specific policy $\theta_i$ according to Eqn ~\ref{eq:3}
        \State Using $\theta$ and $\theta_n$ compute trajectory $\tau_{\theta,\theta_n}$ according to Eqn ~\ref{traj2}
        \EndFor
        \State Compute policy gradient $\nabla_{\theta} L_n(\theta,\theta_n)$ for every agent according to Eqn ~\ref{polgrad} 
        \State Update central policy $\theta \leftarrow \theta + \epsilon \nabla_{\theta} \sum_{n=1}^N  L_{n}(\theta, \theta_n)$ (Eqn ~\ref{eq:7})
      \EndWhile\label{dimapg}
  \end{algorithmic}
\end{algorithm}

\section{Experiments}
\subsection{Environments}\label{sec4.1}
To test the effectiveness of DIMAPG, we perform experiments on both collaborative and competitive tasks. The environments from~\cite{lowe2017multi} and the many-agent (MAgent) environment from~\cite{zheng2017magent} are adapted for our experiments. We setup the following experiments to test out our algorithm :

\begin{figure}[b!]
\centering
\includegraphics[scale =0.3]{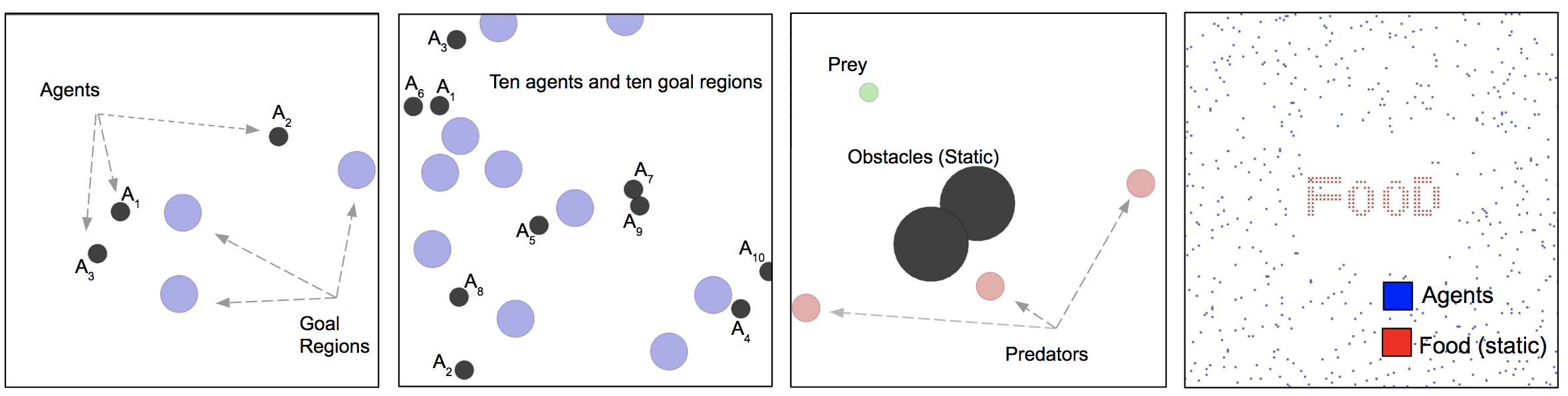}
\caption{\textbf{Multi-agent environments for testing:} We consider both collaborative as well as competitive environments. \textbf{Left:} Cooperative Navigation (with 3 agents) \textbf{Center Left:} Cooperative Navigation for 10 agents. \textbf{Center Right:} Predator-Prey \textbf{Right:} Survival with many (630) agents}
\end{figure}

\textbf{Cooperative Navigation} This task consists of $N$ agents and $N$ goals. All agents are identical, and each agent observes the position of the goals and the other agents relative to its own position. The agents are collectively rewarded based on the how far any agent is from each goal. Further, the agents get negative reward for colliding with other agents. This can be seen as a coverage task where all agents must learn to cover all goals without colliding into each other. We test increasing the number of agents and goal regions and report the minimum reward across all agents.
 
\textbf{Predator Prey} This task environment consists of two populations - predators and preys. Prey are faster than the predators. The environment is also populated with static obstacles that the agents must learn to avoid or use to their advantage. All agents observe relative positions and velocities of other agents and the positions of the static obstacles. Predators are rewarded positively when they collide with the preys and the preys are rewarded are negatively. 

\textbf{Survival} This task consists of a large number of agents operating in an environment with limited resources or food. Agents get reward for eating food but also get reward for killing other agents (reward for eating food is higher). Agents must either rush to get reward from eating food or monopolize the food by killing other agents. However, when the agents kill other agents they incur a small negative reward. Each agent's observations consists of a spatial local view component and a non spatial component. The local view component encodes information about other agents within a range while the non spatial component encodes features such as the agents ID, last action executed, last reward and the relative position of the agent in the environment. 

\subsection{Experimental Results}
For all experiments, we use a neural network policy that consists of two hidden layers with 100 units each and uses ReLU nonlinearity. For the Cooperative Navigation task, we use the vanilla policy gradient or REINFORCE~\cite{reinforce} to compute updates ($\theta_n$) and TRPO~\cite{schulman2015trust} to compute $\theta$. For the Predator Prey and Survival tasks we switch to using REINFORCE for both $\theta$ and $\theta_n$. To establish baselines, we compare against both centralized and decentralized deep MARL approaches. For decentralized learning, we use MADDPG from ~\cite{lowe2017multi} using the online implementation open sourced by the authors. Since the authors in~\cite{lowe2017multi} already show MADDPG agents work better than other methods where individual agents are trained by DDPG, REINFORCE, Actor-Critic, TRPO, DQN, we do not re implement those algorithms. Instead, we implement a centralized A3C (Actor-Critic)~\cite{mnih2016asynchronous} and centralized TRPO that take in as input the joint space of all agents observations and output actions over the joint space of all agents. We call this the \textit{Kitchensink} approach. Details about the policy architecture for \textit{A3C\_Kitchenshink} and \textit{TRPO\_Kitchensink} are provided in the appendix. Our experiments are designed using the rllab benchmark suite~\cite{duan2016benchmarking} and use Tensorflow~\cite{tensorflow2015-whitepaper} to setup the computation graph for the neural network and compute gradients. 
\subsubsection{Cooperative Navigation}
We setup co-operative navigation as described in Section ~\ref{sec4.1}. Agents are rewarded for being close to the goals (negative square of distance to the goals) and get negatively rewarded for colliding into each other or when they step out of the environment boundary. We also observe that in order to stabilize training, we need to clip our rewards in the range [-1,1]. We use a horizon $T=200$ after which episodes are terminated. Additional hyper parameters are provided in the Appendix. 

\begin{wraptable}{r}{4.5cm}
	{\begin{tabular}{|c|c|c|}
	\hline
	\textbf{} & \textbf{\textit{n=3}} &\textbf{n=10}  \\  \hline
	{Using $\theta$}  & -34.8  & -8 \\ \hline
	{Using $\theta_i'$}  & -37.19 & -8.5 \\ \hline
	{Fine Tune} & -44.17 & -56.3 \\ \hline
    \end{tabular}}
\caption{\textbf {Min. reward across all agents after training (avg. over 100 episodes)}} 
\label{table:Results1}
\end{wraptable}

\begin{figure}[t!]
\centering
\includegraphics[scale=0.3]{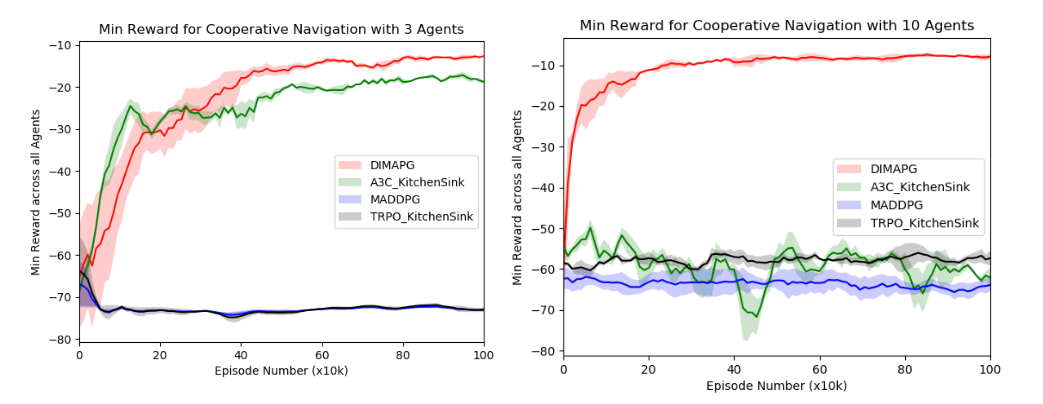}
\caption{\textbf{Min reward vs. number of episodes for Cooperative Navigation:} DIMAPG converges quickly in both scenarios. The protocol followed in the plots involves 5 independent runs for each algorithm with different seeds, darker line represents the mean and the shaded lighter region represents the variance.}
\label{fig:simple_spread}
\end{figure}

We run our proposed algorithm and baselines on  this environment when number of agents $n=3$ and $n=10$. Since the baselines A3C\_Kitchenshink and TRPO\_Kitchensink operate over the joint space, they are setup to maximize the minimum reward across all agents. The training curve for our tasks can be seen in Fig ~\ref{fig:simple_spread}. We notice that for the simple case, A3C\_Kitchenshink performs very well and quickly converges. This is expected since the number of agents is low and the dimensionality of the input space is not large. TRPO\_Kitchenshink and MADDPG perform worse and while they converge, the convergence is only seen after 300-400k episodes. When $n$ is increased to ten, we observe that only DIMAPG is able to quickly learn policies for all agents. 

In our initial hypothesis, we sought to use $\theta$ across all agents since we assumed that the policies for all agents in a given population live close to each other in parameter space. We observe from Table ~\ref{table:Results1} that after training using $\theta$ or $\theta_i'$ (after k-shot adaptation from $\theta$) yields almost similar results thus, verifying our hypothesis. We also consider the case where we train only 1 agent and then run the same policy across all agents. We observe that this yields poor results.

\subsection{Predator Prey}
The goal of this experiment is to compare the effectiveness of DIMAPG on competitive tasks. In this task, there exist 2 populations of agents; predators and preys. Extending our hypothesis to this task, we would like to learn a single policy for all predators and a single policy for all preys. It is important to note that even though, the policies are different, they are trained in parallel which in the centralized setup enables us to condition each agents trajectory on the actions of other agents even if they are in a different population. We experiment with two scenarios; 12vs1 and 3vs1 predator prey games where the prey are faster than the predator. The horizon used is $T=200$. 

\begin{figure}[t!] 
\centering
\includegraphics[scale=0.3]{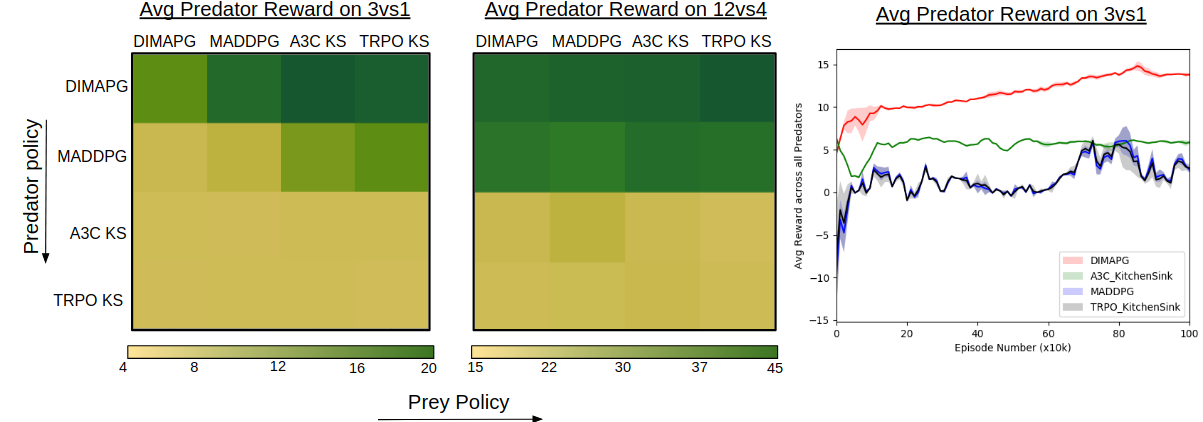}
\caption{\textbf{Results on Predator Prey. Left, Center:} Average predator reward collected over 100 episodes after training different policies for predators and preys. In the 3 Predators vs 1 Prey game, the prey is 30\% faster than the predators. In the 12 Predators vs 4 Prey, the prey is 50\% faster than the predators. \textbf{Right:} Avg predator reward vs episodes during training for 3vs1 game.}
\label{fig:simple_tag}
\end{figure}

Our results are presented in Fig ~\ref{fig:simple_tag}. We observe that DIMAPG is able to effectively learn better policies than both MADDPG and the centralized Kitchensink methods on this competitive task. Similar results with DIMAPG are achieved even when the number of predators and preys are increased. 
\subsection{Survival}
The goal of this experiment is to demonstrate the effectiveness of DIMAPG on environments with a large number of agents. The environment is populated with agents and food (the food is static particles at the center). Agents must learn to survive by eating food. To do so they can either rush to gather food and get reward or monopolize the food by first killing other agents (killing other agents results in a small negative reward). We use DIMAPG to learn the central policy that is deployed across all agents by randomly sampling $N$ agents from the population. We roll out each episode for a horizon of $T=200$. Each environment is populated with $160$ food particles (eating one food particle yields a reward of +5). For this task, it is infeasible to train the other baselines and hence we do not benchmark for this experiment. 

\begin{table}[h!]
\centering
\begin{tabular}{|c|c|c|}
	\hline
	\textbf{Statistics} & \textbf{\textit{N=230}} &\textbf{N=630}  \\  \hline
	{Food Left}  & 0  & 0 \\ \hline
	{Survivors}  & 227 & 490 \\ \hline
	{Average Reward} & 946 & 674\\ \hline
    \end{tabular} 
\caption{\textbf{Statistics on Survival collected over over 100 games using DIMAPG, after training.} Initial average reward for $N=630$ is -3800 and for $N =230$ it is -1530.}\label{tab:survivorstats}
\end{table}
We gauge the performance of DIMAPG on this task by evaluating the number of surviving agents and the food left at the end of the episode as well as the average reward over agents per episode.(Table ~\ref{tab:survivorstats}). It is observed in the case when $N=225$, the agents do not kill each other and instead learn to gather food. When the number of agents is increased to $N=630$ agents close to the food rush in to gather food while those further away start killing other agents. 
\section{Conclusion and Outlook}
Thus, in this work we have proposed a distributed optimization setup for multi-agent reinforcement learning that learns to combine information from all agents into a single policy that works well for large populations. We show that our proposed algorithm performs better than other state of the art deep multi agent reinforcement learning algorithms when the number of agents are increased.

One bottleneck in our work is the significant computation cost involved in computing the second derivatives for the gradient updates. Due to this, in practice we make approximations for the second derivative and are restricted to simple feedforward neural networks. On more challenging tasks, it might be a good idea to try recurrent neural networks and investigate methods such as the one presented in ~\cite{martens2018kroneckerfactored} to compute fast gradients. We leave this for future work.
\clearpage
\small
\bibliographystyle{ieeetr}
\bibliography{nipscite_2018}
\clearpage
\appendix
\textbf{APPENDIX} 
\section{Derivation for Multi-Agent Policy Gradient}
Following Section 4.1, the overall optimization problem for distributed meta-learning was given as :
\begin{equation}
\label{appeq:8}
\max_{\theta} \mathbb{E}_{n \sim P(Pop)} \Big[ \mathbb{E}_{\tau_{n}^\theta \sim P_{n}(\tau^{\theta}_{n}|\theta)} \Big[ \mathbb{E}_{\tau_{n}^{\theta,\theta_n} \sim P_{n}(\tau_{n}^{\theta, \theta_n}|\theta,\theta_n)} [L_{n}(\theta,\theta_n)|(\tau_{n}^{\theta},\theta)] \Big] \Big]
\end{equation}
where trajectories $\tau_{n}^\theta$ and 
$\tau_{n}^{\theta,\theta_n}$ are random variables with distributions $P_{n}(\tau_{n}^\theta|\theta)$ and $P_n(\tau_{n}^{\theta,\theta_n}|\theta, \theta_n)$ respectively.
Assuming, we sample N agents, the above Eqn~\ref{appeq:8} can be rewritten as: 
\begin{equation}
\label{appeq:9}
\max_{\theta} \frac{1}{N}\sum_{n=1}^N \Big[ \mathbb{E}_{\tau_{n}^\theta \sim P_{n}(\tau^{\theta}_{n}|\theta)} \Big[ \mathbb{E}_{\tau_{n}^{\theta,\theta_n} \sim P_{n}(\tau_{n}^{\theta, \theta_n}|\theta,\theta_n)} [L_{n}(\theta,\theta_n)|(\tau_{n}^{\theta},\theta)] \Big] \Big]
\end{equation}
Let :
\begin{equation}
\label{appeq:10}
\mathcal{L}_n(\theta,\theta_n) = \Big[ \mathbb{E}_{\tau_{n}^\theta \sim P_{n}(\tau^{\theta}_{n}|\theta)} \Big[ \mathbb{E}_{\tau_{n}^{\theta,\theta_n} \sim P_{n}(\tau_{n}^{\theta, \theta_n}|\theta,\theta_n)} [L_{n}(\theta,\theta_n)|(\tau_{n}^{\theta},\theta)] \Big] \Big]
\end{equation}
Since it is required that we maximize only over theta, we are interested in marginalizing $\theta_n$. Expanding all expectations we can write:
\begin{equation}
\label{eq10}
\mathcal{L}_n(\theta,\theta_n) = \int \int \int L_{n}(\theta,\theta_n) P_n(\tau_n^{\theta,\theta_n}|(\theta,\theta_n))P_n(\theta_n|\theta,\tau_n^{\theta})P_n(\tau_n^{\theta}|\theta)d\tau_n^{\theta}d\tau_n^{\theta,\theta_n}d\theta_n
\end{equation}

Assuming, we use the k gradient steps instead of just one as presented in Eqn 10 in the main paper, this can be rewritten as :
\begin{equation}
\begin{split}
\label{eq11}
\mathcal{L}_n(\theta,\theta_n) = \int L_{n}(\theta,\theta_n) P_n(\tau_n^{\theta,\theta_n}|(\theta,\theta_n))P_n(\theta_n^{[k]}|\theta_n^{[k-1]},\tau_n^{\theta_n^{[k-1]}}) P_n(\theta_n^{[k-1]}|\theta_n^{[k-2]},\tau_n^{\theta_n^{[k-2]}}) \ldots \\ P_n(\theta_n^{[1]}|\theta_n^{[0]},\tau_n^{\theta_n^{[0]}})P_n(\tau_n^{\theta}|\theta)d\tau_n^{\theta}d\tau_n^{\theta,\theta_{n}^{[0]}}d\tau_n^{\theta,\theta_{n}^{[1]}}\ldots d\tau_n^{\theta,\theta_{n}^{[k]}}d\theta_n
\end{split}
\end{equation}
The term  $P_n(\theta_n|\theta,\tau_{n}^{\theta})d\theta_n$ in the above Eqn~\ref{eq10}  can be integrated out if we assume a delta distribution for $P_n(\theta_n|\theta,\tau_n^{\theta})$:
\begin{equation}
P_n(\theta_n|\theta,\tau_n^{\theta})=\delta \bigg(\theta_0 + \alpha_1 \nabla_{\theta_n} 
L_n(\theta,\theta_n)|_{\theta=\theta_0, \theta_n=\theta_0}\bigg)
\end{equation}
A similar observation can be made for the intermediate terms $P_n(\theta_n^{[1]}|\theta_n^{[0]},\tau_n^{\theta_n^{[0]}}),P_n(\theta_n^{[2]}|\theta_n^{[1]},\tau_n^{\theta_n^{[1]}})$,\\$\ldots$, $P_n(\theta_n^{[k]}|\theta_n^{[k-1]},\tau_n^{\theta_n^{[k-1]}})$ in the above Eqn~\ref{eq11}. 
Thus after integrating these terms out (in the above Eqn~\ref{eq10} or~\ref{eq11}, we are left with: 
\begin{equation}
\label{eq:aftermargin}
\mathcal{L}_n(\theta,\theta_n)= \int \int L_{n}(\theta,\theta_n) P_n(\tau_{n}^{\theta,\theta_n}|(\theta,\theta_n))P_n(\tau_n^{\theta}|\theta)d\tau_n^{\theta}d\tau_{n}^{\theta,\theta_n}
\end{equation}
Taking the gradient of this above equation~\ref{eq:aftermargin} and rewriting it as an expectation form we get:
\begin{equation}
\label{polgrad1}
\begin{split}
\nabla_{\theta} \mathcal{L}_n(\theta,\theta_n) =\mathop{\mathbb{E}}_{\tau_{n}^\theta \sim P_{n}(.|\theta),\tau_{n}^{\theta,\theta_n} \sim P_{n} (.|\theta,\theta_n)} \Bigg[ L_{n}(\theta,\theta_n)\nabla_{\theta} \log \pi_{\theta_{n}}(\tau_{n}^{\theta,\theta_n}) + L_{n}(\theta,\theta_n)\nabla_{\theta}\log \pi_{\theta}(\tau_{n}^{\theta})\Bigg] 
\end{split}
\end{equation}

\section{Connection to Meta-Learning}
We observe that there exists a natural connection between our proposed distributed learning and gradient based meta-learning techniques such as the one used in [23,24]. We briefly introduce gradient based meta-learning here and draw connections from our work to that of meta-learning.

\subsection{Model-Agnostic Meta Learning (MAML)}
Consider a series of RL tasks $\mathcal{T}_{i}$ that one would like to learn. Each task can be thought of as a Markov Decision Process (MDP) $\mathcal{M} (S,A,R,\mathcal{P}')$ consisting of observations $s \in S$, actions $a \in A$, a state transition function $\mathcal{P}'(s_{t+1}|s_t,a_t)$ and a reward function $R(s_t,a_t)$. To solve the MDP (for each task), one would like to learn a policy $\pi: s \rightarrow a$ that maximizes the expected sum of rewards over a finite time horizon $H$, $\max_{\pi} [\sum_{t=1}^H R_t (s_t,a_t)]$. Let the policy be represented by some function $f_{\theta}$ where $\theta$ is the initial parameters of the function. 

In MAML [24] the authors show that, it is possible to learn a policy $\pi_{\theta}$ which can be used on a task $\mathcal{T}_i$ to collect a limited number of trajectories $\mathcal{\tau}_{\theta}$ or experience $\mathcal{D}$ and quickly adapt to a task specific policy $\pi_{\theta_i '}$ that minimizes the task specific loss $ L_{\mathcal{T}_i} (\tau_{\theta}) = -\mathbb{E}_{s_t, a_t \sim \tau_{\theta} }  [\sum_{t=1}^H R_t(s_t,a_t)]$. MAML learns task specific policy $\pi_{\theta_i '}$ by taking the gradient of $L_{\mathcal{T}_i} (\tau_{\theta})$ w.r.t $\theta$. This is then followed by collecting new trajectories $\mathcal{\tau}_{\theta_i '}$ or experience set $\mathcal{D}_{i}'$ using $\pi_{\theta_i '}$ in task $\mathcal{T}_i$. $\theta$ is then updated by taking the gradient of $L_{\mathcal{T}_i} (\tau_{\theta_i '})$ w.r.t $\theta$ over all tasks. The update equations for $\theta'$ and $\theta$ are given as:
\begin{equation}
\theta_i':= \theta - \alpha \nabla_{\theta} L_{\mathcal{T}_i} (\tau_{\theta}), \hspace{1cm} \theta:=\theta - \beta \nabla_{\theta} \sum_{\mathcal{T}_i} L_{\mathcal{T}_i} (\tau_{\theta_i '}) 
\end{equation}
where $\alpha$ and $\beta$ are the hyperparameters for step size. Authors in [23] extend MAML to show that one can think about MAML from a probabilistic perspective where all tasks, trajectories and policies can be thought as random variables and $\theta '$ is generated from some conditional distribution $P(\theta'| \theta, \tau_{\theta})$. 

\subsection{Distributed Optimization for Multi Agent systems}
We observe the meta-policy $\pi_{\theta}$ that MAML attempts to learn and uses as an initialization point for the different tasks is similar in spirit to the central policy $\theta$ DIMAPG attempts to learn and execute on all agents. In both, approaches $\theta$ captures information across multiple tasks or multiple agents. An important difference between our work and MAML or meta-learning is that during execution (post training) we execute $\theta$ while MAML uses $\theta$ to do a 1-shot adaptation for task $\mathcal{T}_i$ and then executes $\theta_i'$ on $\mathcal{T}_i$.

Another interesting point to note here is the difference in the trajectories $\tau_{\theta_i'}$ that is used by MAML and the trajectory $\tau_{n}^{\theta,\theta_n}$ that is used by DIMAPG to update task or agent specific policy $\theta_i'$ or $\theta_n$. In the distributed optimization for multi-agent setting, due to the non-stationarity, it is absolutely necessary that we ensure the other agents are held constant (to $\theta$) while agent $n$ is optimizing its task specific policy $\theta_n$. MAML has no such requirement.

\section{Experimental Details}
\subsection{A3C KitchenSink and TRPO KitchenSink}
For A3C KitchenSink, we input the agents observation and reshape it into a $n \times m$ matrix. This is then fed into a 2D convolution layer with 16 outputs, Elu activation and a kernel size of 2, stride of 1. The output from this layer is fed into another 2D convolution layer with 32 outputs,Elu activation and a kernel size of 2, stride of 1. The output from this layer is flattened and fed into a fully connected layer with 256 outputs and Elu activation. This is followed by feeding into a LSTM layer with 256 hidden units. The output from the LSTM is then fed into two separate fully connected layers to get the policy estimate and the value function estimate. Actor-critic loss is setup and minimzied using Adam with learning rate 1e-4. For TRPO Kitchensink, we setup similar policy layer and value function layer.
\subsection{DIMAPG}
For this task, we used a neural network policy with two hidden layers with 100 units each. The network uses a ReLU non-linearity. Depending on the experiment we compute agent specific gradient updates using REINFORCE and TRPO for the central policy gradient updates. The baseline is fitted separately at each iteration for all agents sampled from the population. We use the standard linear feature baseline. The learning rate for agent specific policy updates $\alpha_1$=$\alpha_2$=0.01. Learning rate for central policy updates $\epsilon=0.05$. In practice, to adapt $\theta$ to $\theta_n$ we do multiple gradient steps. We observe k=3 (number of gradient steps) is a good choice for most tasks. For both $\theta$ and $\theta_n$ updates, we collect 25 trajectories. 

\subsection{Survivor}
In this experiment, the environment is populated with agents and food particles. The agents must learn to survive by eating food. To do so they can either rush to gather food and get reward or monopolize the food by first killing other agents (killing other agents results in a small negative reward). 
Each agent in this environment also has orientation. The agents can either chose to one of 12 neighboring cells or stay as is, or chose to attack any agent or entity in 8 neighboring cells. Finally the agent can also choose to turn right or left. At every step, the agents receive a "step reward" of -0.01. If the agent dies, its given a reward of -1. If the agent attacks another agent, it receives a penalty of -0.1. However, if it chooses to attack another agent by forming a group it receives an award of 1. The agent also gets a reward of +5 for eating food. 

As stated in the main paper, it is observed that in the case when $N=225$, the agents do not kill each other and instead learn to gather food. When the number of agents is increased to $N=630$ agents close to the food rush in to gather food while those further away start killing other agents. We present a snapshot of the learned policy in Figure 1 and Figure 2.

\begin{figure}[h]
\includegraphics[width=\linewidth]{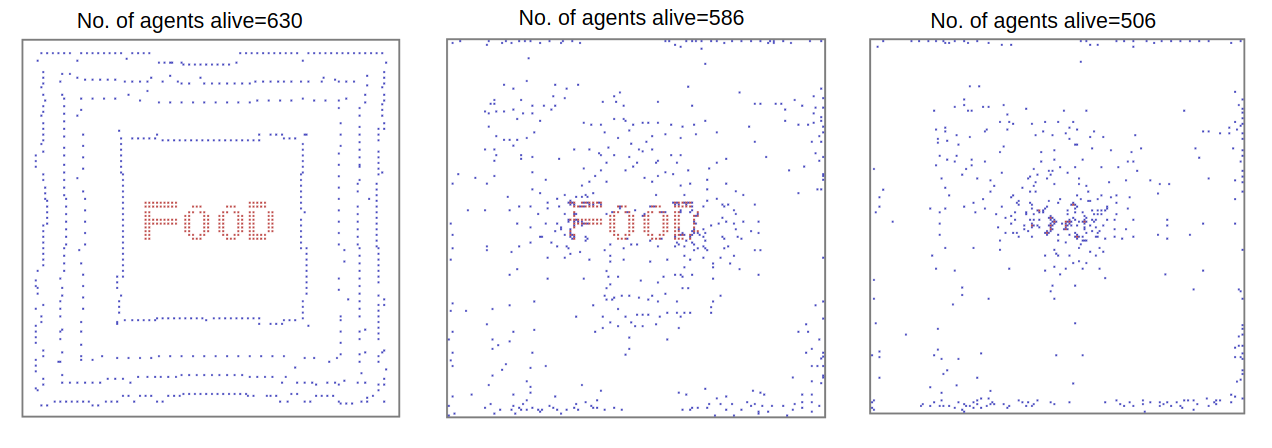}
\caption{\textbf{Learned policy on Survivor(N=230)} When the number of agents is small, agents prefer to eat food instead of killing each other. Most agents survive in this setting.}
\end{figure}
\begin{figure}[h]
\includegraphics[width=\linewidth]{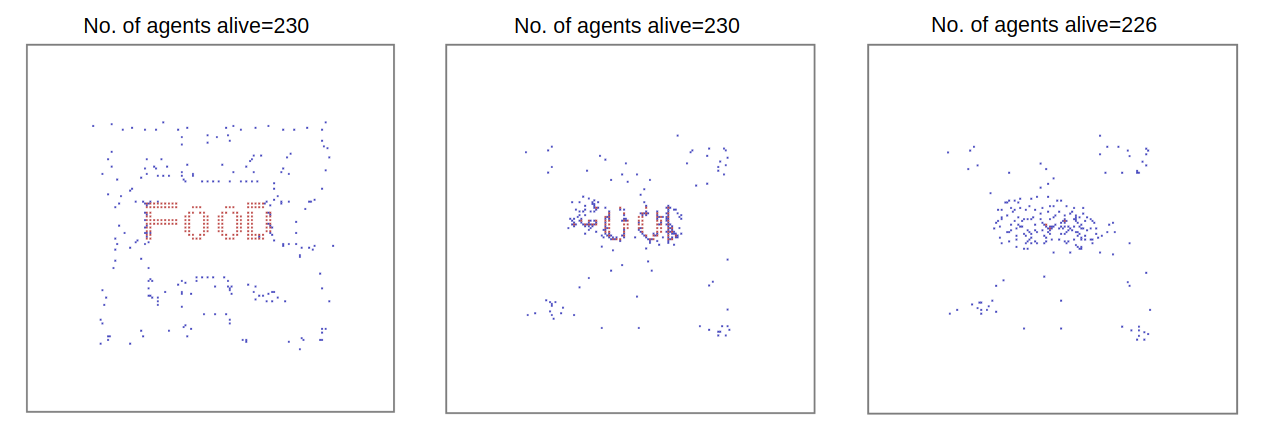}
\caption{\textbf{Learned policy on Survivor(N=630)} When the number of agents is much larger than the amount of food in the environment, the agents closer to the food rush in to gather food. We observe that the agents further away (near the walls) form teams and try to take down other agents thus maximizing reward for the group. This can also be interpreted as follows: Agents who can observe the food within their sensing range choose to rush in food. Agents who do not observe food within their sensing range choose to form groups to take down other agents.}
\end{figure}

\end{document}